\documentclass[sigconf]{acmart}

\usepackage{balance} 
\usepackage[linesnumbered,ruled,vlined]{algorithm2e}
\AtBeginDocument{%
  }

\newcommand{\padex}{\textsc{PADEX}\xspace}

\copyrightyear{2025}
\acmYear{2025}
\setcopyright{rightsretained}
\acmConference[GECCO '25 Companion]{Genetic and Evolutionary Computation Conference}{July 14--18, 2025}{Malaga, Spain}
\acmBooktitle{Genetic and Evolutionary Computation Conference (GECCO '25 Companion), July 14--18, 2025, Malaga, Spain}\acmDOI{10.1145/3712255.3726576}
\acmISBN{979-8-4007-1464-1/2025/07}

\acmSubmissionID{pap208}

\begin{document}

\title{Explainable AI Based Diagnosis of Poisoning Attacks in Evolutionary Swarms}

\author{Mehrdad Asadi}
\affiliation{
  \institution{AI Lab, Vrije Universiteit Brussel}
  \city{Brussels}
  \country{Belgium}}
\email{mehrdad.asadi@vub.be}

\author{Roxana R\u{a}dulescu}
\affiliation{
  \institution{Intelligent Systems, Utrecht University}
  \city{Utrecht}
  \country{Netherlands}}
\email{r.t.radulescu@uu.nl}

\author{Ann Nowé}
\affiliation{
  \institution{AI Lab, Vrije Universiteit Brussel}
  \city{Brussels}
  \country{Belgium}}
\email{ann.nowe@vub.be}

\renewcommand{\shortauthors}{Asadi et al.}


\begin{abstract}
Swarming systems, such as for example multi-drone networks, excel at cooperative tasks like monitoring, surveillance, or disaster assistance in critical environments, where autonomous agents make decentralized decisions in order to fulfill team-level objectives in a robust and efficient manner. Unfortunately, team-level coordinated strategies in the wild are vulnerable to data poisoning attacks, resulting in either inaccurate coordination or adversarial behavior among the agents. To address this challenge, we contribute a framework that investigates the effects of such data poisoning attacks, using explainable AI methods. We model the interaction among agents using evolutionary intelligence, where an optimal coalition strategically emerges to perform coordinated tasks. Then, through a rigorous evaluation, the swarm model is systematically poisoned using data manipulation attacks. We showcase the applicability of explainable AI methods to quantify the effects of poisoning on the team strategy and extract footprint characterizations that enable diagnosing. Our findings indicate that when the model is poisoned above $10\%$, non-optimal strategies resulting in inefficient cooperation can be identified.
\end{abstract}



\maketitle

\section{Introduction}

To enable efficient task execution among a team of agents, it is essential to consider coordinated decision-making rather than individual. To deal with this requirement, swarm intelligence has emerged as a way to optimize the decision-making of self-organizing agents through interactions. However, such a framework is vulnerable to data poisoning attacks, resulting in suboptimal coordination that may cause excessive use of resources of the agents, e.g., induce misbehavior in the swarm to cause targeted damage in safety-critical environments \cite{wang2023prevent}. This vulnerability introduces a significant challenge, particularly as the deployment of autonomous systems becomes more widespread. With the rise of regulations of the EU AI Act \citep{AIAct21}, ensuring reliable and compliant behavior of such systems operating in the wild is important, and therefore developing frameworks and methods to analyze the behavior of AI-driven swarms in the presence of adversaries is highly demanded. 

In this paper, we contribute \padex (Poisoning Attack Diagnosis in Evolutionary swarms with eXplainable AI), \textit{a general diagnosis framework for swarm intelligence} that emerged from the evolutionary behavior of agents. Our framework incorporates the use of machine learning-based surrogate modeling of a black-box evolutionary stable behavior, which helps, together with XAI-based approaches, in identifying anomalous model behavior caused by feature perturbation attacks. To realize such a framework, we provide a concrete instantiation of \padex on a multi-drone cooperative sampling task and show how the existing XAI methods would be able to characterize sub-optimal swarm behavior once targeted by a data poisoning attack. This work lays the groundwork for severity characterization of potential attacks as well as early diagnosis.  

\section{The \padex Diagnosis Framework}
\label{sec::framework1}
\begin{figure}[t!]
    \centering
    \includegraphics[width=0.45\textwidth]{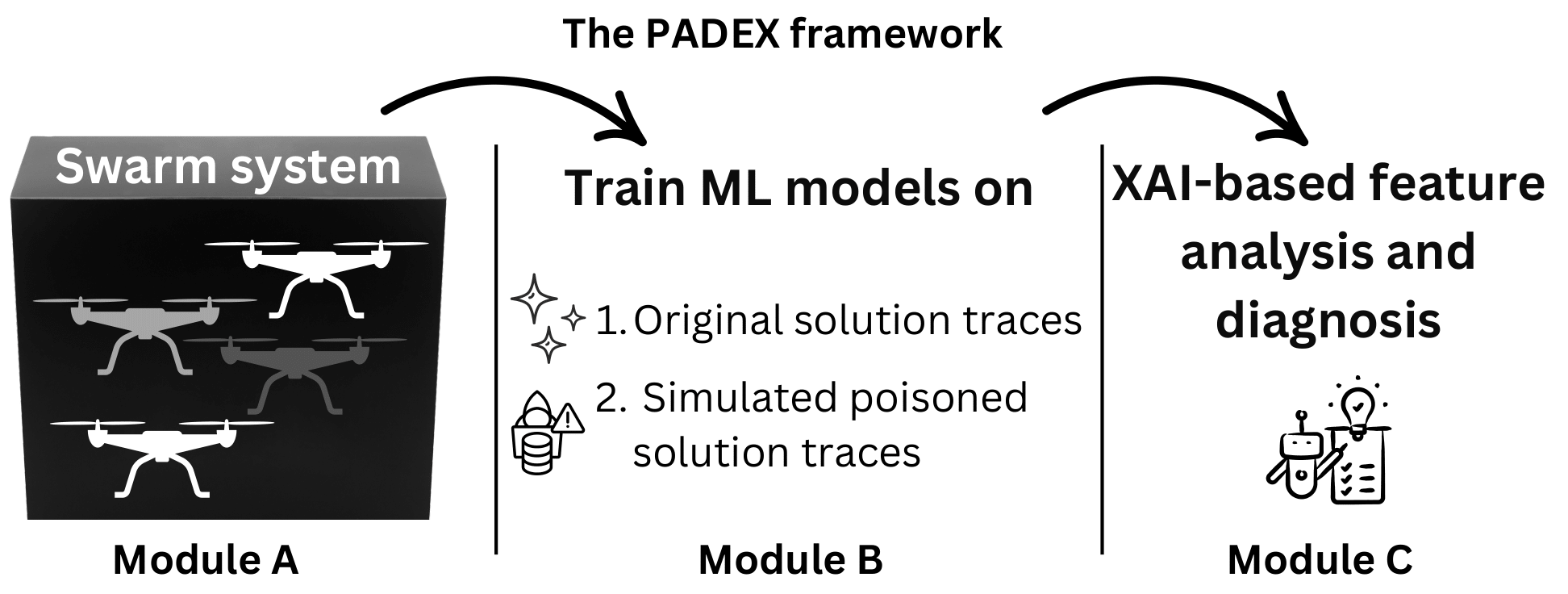}
    \caption{Overview of the modules and pipeline of the \padex framework}
    \label{fig1:general}
\end{figure}

Deploying swarm-based systems in the wild is a challenging endeavor, requiring a careful strategic study of their interactions, to ensure efficiency and robustness~\cite{alqudsi2025exploring}. Furthermore, such infrastructures remain vulnerable to data poisoning attacks after deployment \cite{wang2023prevent}. To this end, we propose \padex, a diagnosis framework that is able to analyze and diagnose the state of the system using generated behavior traces. Figure~\ref{fig1:general} presents a high-level overview of \padex. The general framework can be broken down into three main modules. We provide below a detailed description of each module and their interaction. 

\textbf{\textit{A - Swarming system.}} This black-box module enables flexible implementation of swarm agents solving coordination problems. It supports various swarm-based algorithms or any other bio-inspired methods, allowing seamless integration. By leveraging emergent dynamics from local interactions, it establishes robust collective behavior and a stable solution space to be analyzed thereafter.

\textbf{\textit{B - Surrogate models.}} Before deployment, \padex assumes an evolutionary behavior from the previous component and approximates a benign ML-based surrogate model in Module B. Once a stable fingerprint model is established, deployment can proceed. In Section ~\ref{sec:threat}, we show how these surrogate models, combined with XAI-based methods, detect manipulated or augmented false data by simulating feature perturbations, such as data poisoning attacks.

\textbf{\textit{C - XAI-based diagnosis.}} This module aims to monitor the behavior of a benign swarm model in comparison to a potentially poisoned deployment model. The key insight is that by looking at the important features used to train the model, it is then possible to identify abnormal emergent behavior.

In the followings section, we present a concrete and complete instantiation of our framework. We then use this instantiation to carry out a rigorous empirical evaluation and validation. 

\section{\padex Instantiation}

\begin{figure*}
    \centering
    \includegraphics[width=0.99\textwidth]{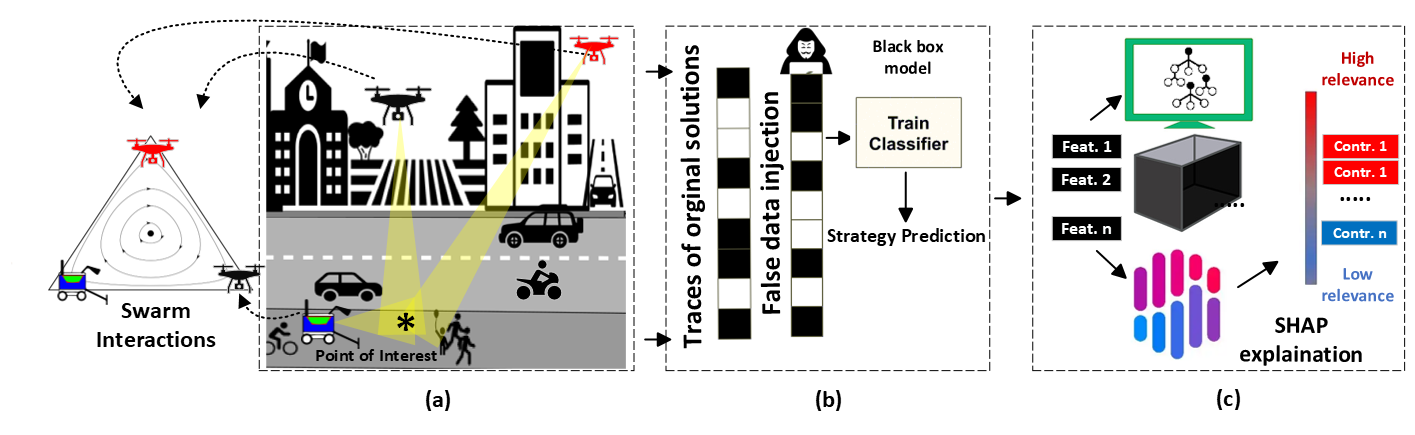}
    \vspace{-0.25cm}
    \caption{An instantiation of \padex; a) Evolutionary Game Modeling b) Feature Perturbation and Model Training (Random Forest); and c) SHAP-based diagnosis}
    \label{fig1:overview}
\end{figure*}

An overview of our framework instantiation is shown in Figure~\ref{fig1:overview} which consists of three steps. In the following, we briefly describe each step of our framework. We consider the task of cooperative sampling \citep{wang2023core}, a wide-encountered problem in monitoring or surveillance scenarios. Section~\ref{sec:modA} describes the problem modeling and employed solution, namely a coalition formation evolutionary-based approach. Sections~\ref{sec:threat} and ~\ref{sec:modC} then describe the benign surrogate model training, attack simulation, and how we propose to use XAI-based methods for analysis and diagnosis (i.e., SHAP~\cite{lundberg2017unified}). 

\subsection{Evolutionary Dynamics of Swarm} 
\label{sec:modA}
We consider a swarm using evolutionary intelligence to form optimal coalitions for cooperative sampling. The goal is to measure Points of Interest (PoI) efficiently by optimizing coalition composition. Agents choose strategies for cooperation, but the game is non-monotone, as adding more agents can introduce overhead without increasing profit. Payoffs depend on each member's contribution, and we model the coalition formation problem using a strategic-form game.
We adopt an evolutionary game approach as an alternative to classical Nash equilibrium analysis. Agents follow inertia and myopia properties based on population games~\cite{sandholm2010population} and behavioral economics~\cite{gal2006psychological}. Initially, players act randomly and adjust strategies via a revision protocol (e.g., replicator dynamics) based on profit. To encourage diversity, we penalize coalitions with high cosine similarity and reward those with members from diverse angular distances. 
\subsection{Assumptions and Threat Modeling}
\label{sec:threat}
As highlighted above, we follow a data-driven approach by applying ML to predict evolutionary strategies to efficiently predict the strategic decisions made by agents in various evolutionary scenarios. Therefore, in this setting, we consider a data poisoning setup where an attacker can inject false location data (e.g., feature perturbation) to poison the overall model~\cite{zhang2022multi}. These type of attacks can be easily performed in this setting without physical access to agents ~\cite{chen2017false}. Our ML model just requires the location coordinates of each agent to estimate a cooperative strategy. Thus, we assume that GPS coordinates can be falsified using fake GPS applications installed on the agent as an exploit~\cite{noh2019tractor}. We assume that all swarm members can see the PoI, though each has a different view. The sampling quality depends on the distance and potential blind spots caused by other swarm members participating in the task. As illustrated in Figure \ref{fig1:overview}(a), a group of autonomous agents (e.g., drones) cooperate to capture valuable data from the PoI effectively. The goal is to perform coordinated sampling of the PoI with the lowest overhead on swarm members. 

\padex assumes the existence of a benign swarm model. Once such a model exists, we inject a set of malicious data into the training process (Figure \ref{fig1:overview}(b)). This can be easily performed by injecting fake GPS coordinates and sub-optimal coalitions. In our experiments, data poisoning is applied at different levels to assess the effect of attack in swarm models. 
\subsection{Feature Importance Analysis}
\label{sec:modC}
Figure \ref{fig1:overview}(c) shows the overall XAI analysis. The key insight is that by looking at the important features during inference, it is then possible to trace back to the benign model and identify anomalous behavior in which coalitions are formed, indicating that optimal coordination has been altered. 

To characterize and quantify the poisoning of the swarm model, we rely on a well-known explainability method to analyze the black-box generated models. We selected SHAP~\cite{lundberg2017unified} as a unified framework for interpreting AI predictions. SHAP assigns each feature an important value for a particular prediction, and it takes advantage of Shapley values, a widely used approach from cooperative game theory. By looking at SHAP deviations and establishing quantifiable differences, it is possible to assess the decision-making abnormalities. 
\section{Experimental Setup}
We solve the proposed game using the evolutionary game approach and to better approximate the emerged behaviour, we generate a dataset of \textit{10,000} solutions by solving the game for each initial random configuration, enabling us to build an ML model (Figure \ref{fig1:overview}(b)) for early prediction of strategies. For this problem instance, we used Random Forest to predict optimal coalitions, resulting in an accuracy of $90\%$. Once our benign swarm model is trained within the space of valid solutions, we capture the fingerprint of feature interactions. We then proceed to augment the training data with poisoned data where we synthetically generate data with altered features. We then apply incremental poison levels to the data. We induce 10\% to 40\% (in 5\% increments) to increase attack severity gradually. We did not consider a higher percentage of poisoned data since capturing its effect is not noticeable anymore as the poisoning data takes over the inference process. 
Finally, we apply the SHAP method to quantify and characterize the deviations via SHAP values used in the inference process to compare the increasing effect of poisoning in our model.
\section{Results} \label{sec::results}
Our key findings include: \textit{(1)} Poison attacks can affect the stable space of solutions of optimal strategy predictions and cause the model to form non-stable strategy that break the self-enforcement rules; \textit{(2)} In addition to reducing the swarm model's accuracy, our results suggest that the deviation of swarm members towards the non-stable starts occurring more evidently when the poisoning reaches a level beyond 10\%; (3) SHAP values can quantify characterizable deviations caused by poisoning, suggesting that poisoning attacks can be detected when comparing clean and poisoned versions of the model.

Besides causing performance degradation (an accuracy drop from $91\%$ to $63\%$) after testing the AI model compromised by poisoning attacks, we can observe that the strategy predictions for the test data change, which can lead to a sub-optimal strategy and coordination. Indeed, the poisoning attack caused the swarm to form inefficient coalitions with a higher cost. This variation in solution distribution is evidence of deviating participants toward forming non-optimized coalitions. Thus, besides providing sub-optimal coordination to each individual agent, our result indicates that the cost associated with forming a coalition also increases.



\begin{figure}[t]
    \centering\includegraphics[width=0.7\columnwidth]{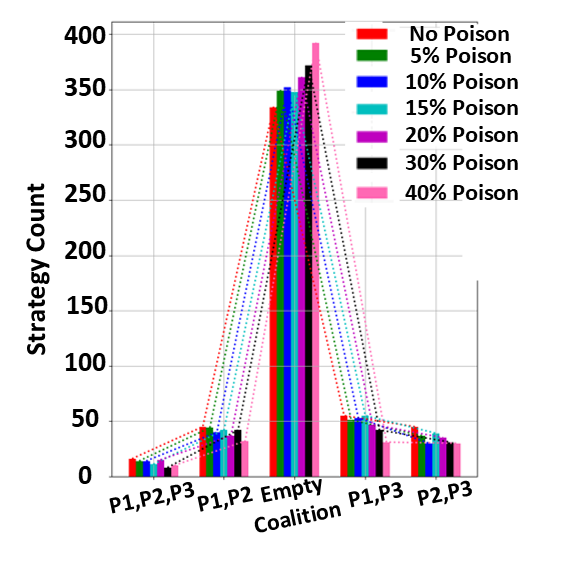}
    \caption{Distribution of predicted strategies for the testing data}
    \label{fig:test3}
\end{figure}
\begin{figure}
    \centering
    \includegraphics[width=\columnwidth]{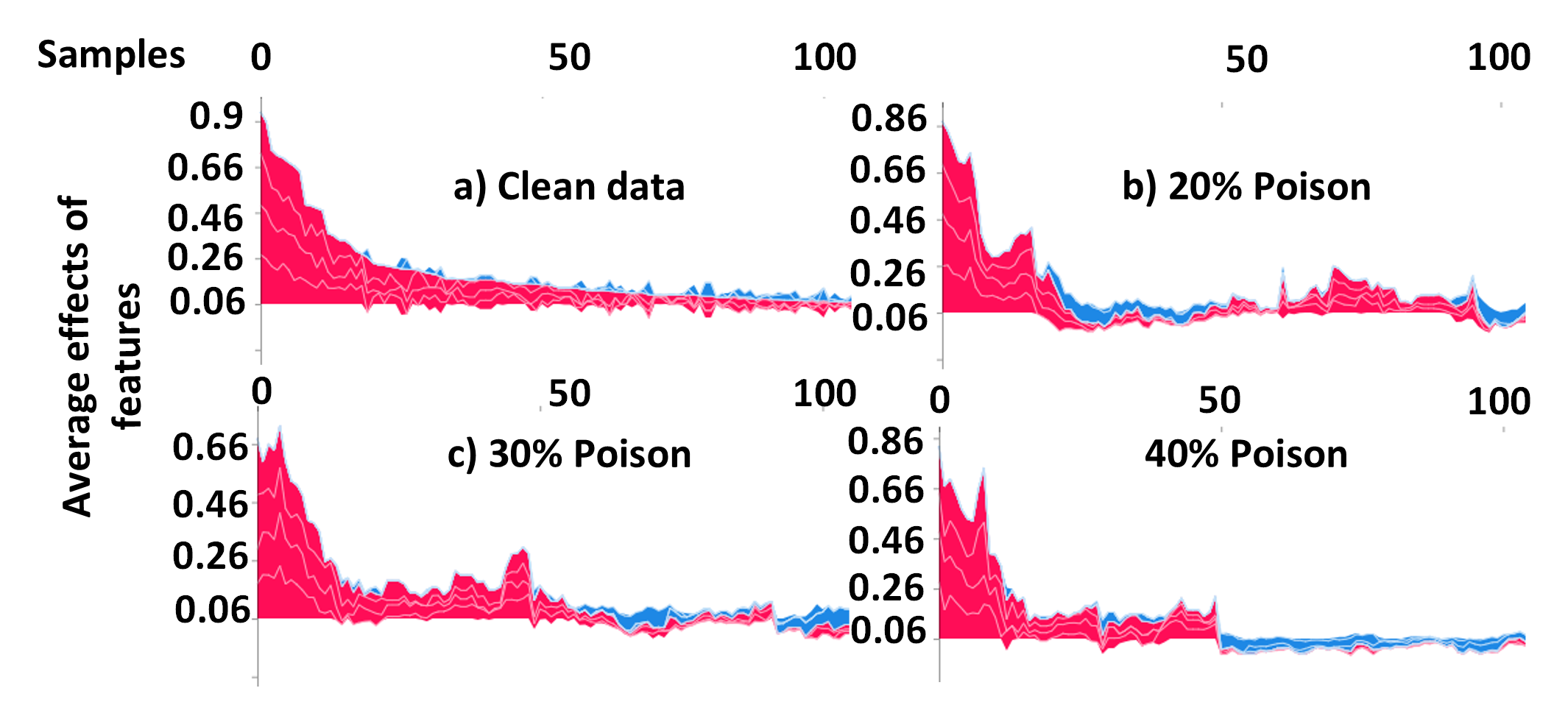}
    \caption{Effects of features on the output of 100 samples. The features are agents' positions (x,y) in a grid-like environment. Blue band shows positive contribution, and pink color shows negative impact on prediction.}
    \label{fig:featureeffect}
\end{figure}


\subsection{\textbf{XAI-based Feature Quantification}}
We next employ SHAP method to analyze the model's behavior after poisoned data has been fed into the training phase. We rely on the generated traces of game solutions\footnote{\url{https://github.com/mehrdadasadiut/GameSolutions-Data}} as it contains many records that enable us to analyze the XAI method comprehensively. 

Quantifying feature importance using SHAP, we can observe that the effects of the poisoning attack up to $10\%$ affect the value of agents' contributions toward the model output in a negative manner. However, as the poisoning severity increases, the level of contribution from different agents changes significantly and causes the contribution of agents toward game outcomes to change, not satisfying the expected observation; therefore, model performance decreases to $63\%$ at a poisoned level of $40\%$. This result is in line with \textit{average effects of features analysis}. As depicted in Figure~\ref{fig:featureeffect}, we plotted the average effects of features for 100 samples for various levels of false data injection. Mann-Whitney-U-Test for each attack severity shows a significant difference (Clean vs. $20\%$ Poison $U=1373.00$, p-value$=0.2574 > 0.05$  - Clean vs. $30\%$ Poison $U=968.50$, p-value$=0.0005 < 0.05$ - Clean vs. $40\%$ Poison $U=982.50$, p-value$=0.0007 < 0.05$) between the average effects of features throughout 100 samples. The result indicates that XAI is an advisable metric for sample-efficient poisoning attack detection (e.g., 100 test samples) when comparing the outcomes against a perfect model.

\section{Related Work}
Evolutionary Computation (EC) is widely used for solving complex multi-agent coordination problems but faces challenges due to its computational intensity. To improve efficiency, researchers integrate ML techniques to approximate fitness evaluations, reducing computational costs~\cite{jin2011surrogate}. Surrogate models like regression, neural networks, and Gaussian processes help bypass expensive computations. For instance, \citet{tzruia2023fitness} proposed a method for fitness approximation through ML, demonstrating that ML methods can effectively replace direct fitness evaluations in genetic algorithms.

On a parallel note, understanding the internal logic of ML models is critical to foster the trustworthiness of systems and applications to users~\cite{glikson2020human}. Multiple XAI methods have been developed to expose the internal logic of models, e.g., SHAP \citep{lundberg2017unified}, LIME \cite{ribeiro2016should}, and occlusion sensitivity \citep{zeiler2014visualizing}, to mention a few. Recently, \citet{ottun2022toward} also evaluated model robustness through the lens of XAI methods, leveraging feature importance to reveal model weaknesses and decision inconsistencies under adversarial image conditions, in an image classification task. In the context of autonomous drones, XAI methods have been utilized to analyze AI models running in individual autonomous drones~\cite{mankodiya2021xai}. While significant progress has been made in integrating ML with evolutionary computation, one emerging research challenge is the study of the impact of data manipulation on the learned and approximated models~\cite{goodfellow2014explaining}. Inspired by  \cite{ottun2022toward}, we generalize and re-purpose the approach in the context of evolutionary swarm systems and attack diagnosis. Our study presents the \padex diagnosis framework to characterize the swarm's emerged misbehavior targeted by feature perturbation attacks. These insights suggest that XAI not only aids in diagnosing model vulnerabilities but also provides a foundation for developing more resilient learning frameworks.
\section{Conclusion and Future Work}
In this work, we presented \padex, a framework designed to detect when swarm intelligence models are compromised by data poisoning attacks. By employing an evolutionary game theoretic approach, we modeled a space of stable strategies for coalition formation. Through the use of SHAP values, we analyzed deviations in coalition predictions, enabling the detection of compromised models. Our extensive experiments demonstrate that our framework effectively identifies quantifiable changes in key model features, signaling when the swarm models are disrupted by attacks. 

In future work, we plan to extend this approach to analyze other attack vectors, such as label flipping in federated learning environments. Our contribution lays the groundwork for methods that not only assess attack severity but also recognize specific attack types, allowing for early mitigation of their effects.
\begin{acks}
Mehrdad Asadi and Ann Nowé were supported by the Flemish AI Research Program and PEER project (EU Horizon Grant 101120406). Roxana R\u{a}dulescu was partly supported by FWO (Grant 1286223N). We thank Huber Flores and Farooq Dar for their technical discussions on the topic.
\end{acks}

\bibliographystyle{ACM-Reference-Format}


\end{document}